# Hypergraph: A Unified and Uniform Definition with Application to Chemical Hypergraph and More

Daniel T. Chang (张遵)

*IBM (Retired)* dtchang43@gmail.com

**Abstract:** The conventional definition of hypergraph has two major issues: (1) there is not a standard definition of directed hypergraph and (2) there is not a formal definition of nested hypergraph. To resolve these issues, we propose a new definition of hypergraph that unifies the concepts of undirected, directed and nested hypergraphs, and that is uniform in using hyperedge as a single construct for representing high-order correlations among things, i.e., nodes and hyperedges. Specifically, we define a hyperedge to be a simple hyperedge, a nesting hyperedge, or a directed hyperedge. With this new definition, a hypergraph is nested if it has nesting hyperedge(s), and is directed if it has directed hyperedge(s). Otherwise, a hypergraph is a simple hypergraph. The uniformity and power of this new definition, with visualization, should facilitate the use of hypergraph for representing (hierarchical) high-order correlations in general and chemical systems in particular. Graph has been widely used as a mathematical structure for machine learning on molecular structures and 3D molecular geometries. However, graph has a major limitation: it can represent only pairwise correlations between nodes. Hypergraph extends graph with high-order correlations among nodes. This extension is significant or essential for machine learning on chemical systems. For molecules, this is significant as it allows the direct, explicit representation of multicenter bonds and molecular substructures. For chemical reactions, this is essential since most chemical reactions involve multiple participants. We propose the use of chemical hypergraph, a multilevel hypergraph with simple, nesting and directed hyperedges, as a single mathematical structure for representing chemical systems. We apply the new definition of hypergraph, with visualization, to chemical hypergraph and, as simplified versions, molecular hypergraph and chemical reaction hypergraph.

# 1 Introduction

*Graph* has been widely used as a mathematical structure for machine learning on molecular structures [1-2] and 3D molecular geometries [3]. However, graph has a *major limitation*: it can represent only *pairwise correlations* between nodes. *Hypergraph* extends graph with *high-order correlations* among nodes represented as *hyperedges*.

This extension is significant or essential for machine learning on *chemical systems* [9]. For *molecules*, this is significant as it allows the direct, explicit representation of *multicenter bonds* and *molecular substructures* (e.g., rings, delocalized bonds, conjugated bonds and functional groups) as high-order hyperedges. For *chemical reactions*, this is essential since most chemical reactions involve *multiple participants (as reactant, product or catalyst)* which can be represented as high-order hyperedges but not pairwise edges.

We propose the use of *chemical hypergraph*, a *multilevel hypergraph* with simple, nesting and directed hyperedges, as a single mathematical structure for representing *chemical systems* consisting of molecules, their constituent atoms and chemical bonds, and their chemical reactions.

The conventional definition of hypergraph [4-8], however, has *two major issues*: (1) there is *not a standard definition of directed hypergraph*, and the notion of direction may be applied either between hyperedges [10=11] or between node sets within the same hyperedge [12-14], and (2) there is *not a formal definition of nested hypergraph*, in which hyperedges can contain other hyperedges.

To resolve these issues, which is required by chemical hypergraph, we propose *a new definition of hypergraph* that *unifies* the concepts of undirected, directed and nested hypergraphs, and that is *uniform* in using hyperedge as a single construct for representing high-order correlations among things, i.e., nodes and hyperedges. Specifically, we define a *hyperedge* to be a *simple hyperedge* (a node set), a *nesting hyperedge* (a hyperedge set), or a *directed hyperedge* (a nesting hyperedge with the notion of direction applied between member hyperedges). With this new definition, a hypergraph is *nested* if it has nesting hyperedge(s), and is *directed* if it has directed hyperedge(s). Otherwise, a hypergraph is a *simple* hypergraph, unnested and undirected.

We apply the new definition of hypergraph, with visualization [15], to *chemical hypergraph* (as multilevel hypergraph) and, as simplified versions, *molecular hypergraph* (as simple hypergraph) and *chemical reaction hypergraph* (as directed hypergraph).

## 2 Hypergraph

In a *hypergraph*, a *hyperedge* can connect any number of *nodes*, unlike an ordinary graph where an edge connects exactly two nodes. Hypergraphs offer a rich mathematical framework that extends beyond ordinary graphs, allowing us to explore *high-order correlations* among nodes in a more flexible manner.

### 2.1 Background: Conventional Definition of Hypergraph

Formally, a *hypergraph* is defined as [4-8]

$G = (V, E, \mathbf{X}, \mathbf{U})$,

where

–

- V is a set of *n nodes (vertices)* $\{v_1, \ldots, v_n\}$,
- E is a set of *m hyperedges*, with each hyperedge $e = \{v_1, \ldots, v_{|e|}\}$ being a subset of V with *order* $|e|$,
- **X** $\in R^{n \times d}$ are *node features* of d dimensions, and
- **U** $\in R^{m \times d'}$ are *hyperedge features* of d' dimensions.

A hypergraph G is thus an ordered pair (V, E) where V is a *finite set* and E is a *multi-set* consisting of multi-subsets of V (a multi-subset is a subset of a set where the elements are allowed to appear more than once). The *order* of a hypergraph is the number of nodes (n), and the *size* is the number of hyperedges (m).

For many applications, it is sufficient and advantageous to use a single *hyperedge feature: weight*. In which case, a *hypergraph* is commonly defined as [4-5]

G = (V, E, **X**, **W**),

where **W** is the diagonal *hyperedge weight matrix*. A further simplification is to treat all hyperedges with equal weight:

G = (V, E, **X**).

**Substructure and Sub-hypergraph**

A *substructure* is the node set of a hyperedge. A *sub-hypergraph* of a given hypergraph is one whose node set and hyperedge set are, respectively, the subset of that of the given hypergraph. A substructure is, therefore, a sub-hypergraph with a single hyperedge.

**Incidence Matrix**

The *incidence matrix* of a hypergraph is a $n \times m$ matrix **H** where rows represent the nodes and columns represent the hyperedges, such that:

**H**(v, e) = 1 if v $\in$ e, and 0 otherwise.

**Directed Hypergraph**

A hypergraph can be *undirected* (usually) or *directed*. For *directed hypergraph*, there is *not a standard definition* [5], and the notion of *direction* may be applied either between hyperedges [10-11] or between node sets within the same hyperedge [12-14]. In either case, the node set in a *directed hyperedge* can be further divided into two sets: the *source (tail) node set* and the *target (head) node set*.

On directed hypergraph, the *incidence matrix* is defined as follows:

$H(v, e) = -1$ if $v \in S(e)$, $1$ if $v \in T(e)$, and $0$ otherwise,

where $S(e)$ and $T(e)$ are the source and target node set for hyperedge $e$, respectively. The incidence matrix $H$ is split into two matrices, $H_s$ and $H_t$, describing the source and target nodes for all hyperedges, respectively.

Directed hypergraphs offer a way to represent complex relationships where multiple elements can *influence* multiple other elements in a specific *direction*. They find applications in various fields like modeling chemical reactions, database modeling, and social network analysis.

**Nested Hypergraph**

A *nested hypergraph* extends the concept of a regular hypergraph by *allowing hyperedges to contain other hyperedges*. In a nested hypergraph, some hyperedges can entirely contain other hyperedges, *forming a hierarchical structure*. These inner hyperedges are nested within the outer hyperedge. This nesting can happen *multiple levels deep*, creating complex relationships between elements.

There is *not a formal definition* of nested hypergraph that we are aware of. The implicit definition, commonly used, is that a hyperedge $e_1$ is *nested* in a hyperedge $e_2$ if the node set of $e_1$ is a subset of the node set of $e_2$.

Nested hypergraphs can model systems with inherent *hierarchical structures*, and can be useful in various fields where hierarchical relationships and interactions between elements play a crucial role, e.g., modeling biological pathways, social network analysis, and named entity recognition.

## 2.2 Hypergraph: Unified and Uniform Definition

As mentioned previously, the conventional definition of hypergraph has two *major issues*: (1) there is not a standard definition of directed hypergraph, and the notion of direction may be applied either between hyperedges or between node sets within the same hyperedge, and (2) there is not a formal definition of nested hypergraph, in which hyperedges can contain other hyperedges. To resolve these issues, which is required by chemical hypergraph, we propose a *new definition of hypergraph* that *unifies* the concepts of undirected, directed and nested hypergraphs, and that is *uniform* in using hyperedge as a single construct for representing high-order correlations among things, i.e., nodes and hyperedges.

We define a hyperedge as:

$G = (V, E, \mathbf{X}, \mathbf{U})$.

where

- V is a set of *n nodes (vertices)* $\{v_1, \ldots, v_n\}$,
- E is a set of *m hyperedges*,
- $\mathbf{X} \in \mathbb{R}^{n \times d}$ are *node features* of d dimensions, and
- $\mathbf{U} \in \mathbb{R}^{m \times d'}$ are *hyperedge features* of d' dimensions.

The difference from the conventional definition of hypergraph comes in our *new definition of hyperedge*:

- $e = e_S \mid e_N \mid e_D$, with
- $e_S = \{v_1, \ldots, v_{|e_S|}\}$ being a subset of V with *order* $|e_S|$,
- $e_N = \{e_1, \ldots, e_{|e_N|}\}$ being a set of hyperedges with order $|e_N|$, and
- $e_D = (e_1, e_2)$,

where $e_S$ denotes *simple hyperedge*, $e_N$ denotes *nesting hyperedge*, and $e_D$ denotes *directed hyperedge*. $e_D$ is an *ordered set* with $e_1$ as the *source hyperedge* and $e_2$ as the *target hyperedge*.

To paraphrase, a hyperedge is a simple hyperedge (i.e., a node set), a nesting hyperedge (i.e., a hyperedge set), or a directed hyperedge (i.e., a nesting hyperedge with the notion of direction applied between member hyperedges). This is *visualized* below (solid circle: node, rounded rectangle: hyperedge):

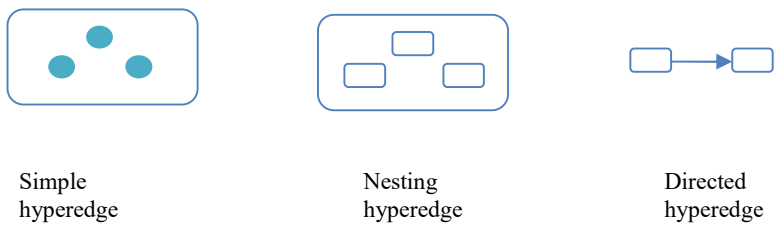

| Simple hyperedge | Nesting hyperedge | Directed hyperedge |

With the new definition, a hypergraph is *nested* if it has nesting hyperedge(s), and is *directed* if it has directed hyperedge(s). Otherwise, a hypergraph is a *simple* hypergraph, unnested and undirected. The uniformity and power of the new definition, with visualization, should facilitate the use of hypergraph for representing (hierarchical) high-order correlations in general and chemical systems in particular.

### Nesting Hyperedge Reduction and Directed Hyperedge Expansion

When needed, a nesting hyperedge with one member can be reduced to the member itself, i.e., {e} => e:

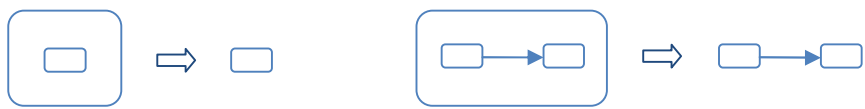

Also, when needed, a directed hyperedge with nested directed hyperedge(s) can be expanded to a series of directed hyperedges, e.g., $((e_1, e_2), (e_3, e_4)) \Rightarrow (e_1, e_2, e_3, e_4)$:

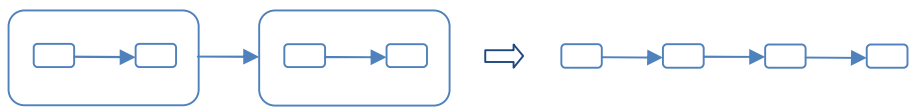

### Characteristics

The new definition has the following characteristics:

- *Unified*: It unifies the concepts of undirected, directed and nested hypergraphs.

- *Uniform*: It uses hyperedge as a single construct for representing high-order (and pairwise) correlations among things, i.e., nodes and hyperedges.
- *Sound*: It is based on set theory and order theory for sets and ordered sets, respectively.
- *Visual*: It has corresponding visual notations for simple hyperedge, nesting hyperedge and directed hyperedge. The notations are consistent with the notions of sets and ordered sets.

## 3 Chemical Hypergraph

*Graph* has been widely used as a mathematical structure for machine learning on molecular structures [1-2] and 3D molecular geometries [3]. However, graph has a *major limitation*: it can represent only *pairwise correlations* between nodes. *Hypergraph* extends graph with *high-order correlations* among nodes represented as *hyperedges*. This extension is significant or essential for machine learning on *chemical systems* [9].

For *molecules*, the extension is significant as it allows the direct, explicit representation of *multicenter bonds* and *molecular substructures* as high-order hyperedges. Some of the most common molecular substructures are:

- *Aliphatic chains*: These are unbranched or branched chains of carbon atoms that are connected by *single bonds*.
- *Conjugated bonds*. These arise when *multiple double bonds are separated by single bonds* within a molecule. Conjugated systems can be *linear, cyclic, or a combination of both*.
- *Rings*: These are closed loops of atoms that are covalently bonded together. Rings can be *aromatic* (having alternating single and double bonds) or *aliphatic* (having only single bonds).
- *Aromatic hydrocarbons*: These are hydrocarbons that contain one or more aromatic rings.
- *Aromatic heterocycles*: These are aromatic rings that contain one or more *atoms other than carbon*, such as nitrogen, oxygen, or sulfur.
- *Functional groups*: These are specific arrangements of atoms within a molecule that are responsible for the molecule's characteristic *chemical behavior*. Some common functional groups include: alcohols (-OH), carboxylic acids (-COOH). amines (-NH2), alkenes (C=C), alkynes (C≡C), aldehydes (-CHO) and Ketones (>C=O).

For *chemical reactions*, the extension is essential since most chemical reactions involve *multiple participants (as reactant, product or catalyst)* which can be represented as high-order hyperedges but not pairwise edges. For example, traditionally the reaction A + B → C + D is modeled as the directed graph whose directed edges corresponded to A → C, A

→ D, B → C and B → D [9]. The problem of this representation is that it introduces artifacts (a set of 9 possible reactions) that hinder the recovery of the original reaction.

We propose the use *chemical hypergraph*, a *multilevel hypergraph* (with simple, nesting and directed hyperedges), as a single mathematical structure for representing chemical systems consisting of molecules, their constituent atoms and chemical bonds, and their chemical reactions.

## 3.1 Chemical Hypergraph as Multilevel Hypergraph

A chemical hypergraph is a *multilevel hypergraph* with simple, nesting and directed hyperedges. *Nodes* in a chemical hypergraph represent *atoms*, and the multilevel hyperedges represent *chemical bonds / molecular substructures, molecules,* and *chemical reactions*, respectively from bottom to top. The *chemical bond / molecular substructure hyperedge* captures the fact that a chemical bond or molecular structure is formed from multiple atoms; the *molecule hyperedge* captures the fact that a molecule consists of multiple chemical bonds and molecular substructures; the *chemical reaction hyperedge* captures the fact that a chemical reaction involves multiple molecules interacting at once;.

To summarize, the *key features* of chemical hypergraphs are:

- *Nodes*: Each node represents an *atom* in a chemical system, with features such as atom type.
- *Bond / Substructure Hyperedges*: Each *simple hyperedge* represents a chemical bond or molecular substructure among atoms, with features such as bond / substructure type. The hyperedge is *nested* within a molecule hyperedge because the chemical bond or molecular substructure always exists as part of a molecule.
- *Molecule Hyperedges*: Each *nesting hyperedge* represents a molecule which consists of multiple chemical bonds and molecular substructures. The hyperedge is *nested* within a chemical reaction hyperedge if the molecule is involved in a chemical reaction.
- *Chemical Reaction Hyperedges*: Each *directed hyperedge* represents a chemical reaction which involves multiple molecules. The hyperedge specifies the *source* and *target hyperedges,* which respectively represent the *reactant* and *product molecules* of the chemical reaction.

This is *visualized* below:

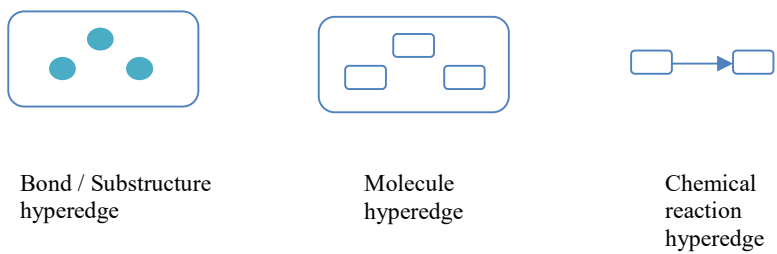

| Bond / Substructure hyperedge | Molecule hyperedge | Chemical reaction hyperedge |

As an example, for the chemical system consisting of the chemical reaction $C_6H_6 + 3H_2 \rightarrow C_6H_{12}$, its reactant and product molecules, and the molecules' constituent chemical bonds and atoms, the chemical hypergraph is visualized below:

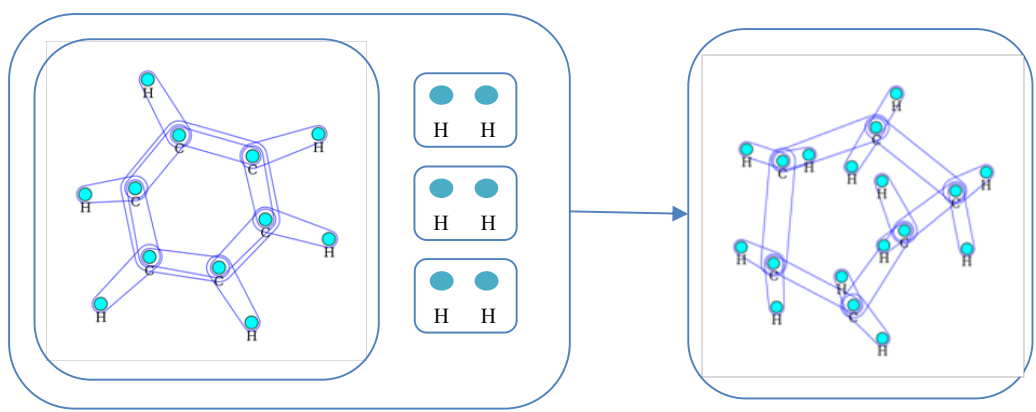

Chemical hypergraph: $C_6H_6 + 3H_2 \rightarrow C_6H_{12}$

As can be seen, *chemical hypergraph* provides *a uniform, powerful and flexible representation of chemical systems* (from atoms, to chemical bonds / molecular substructures, to molecules, to chemical reactions) for machine learning and other usage.

**Characteristics**

Chemical hypergraph has the following characteristics:

- *Complete*: All essential elements in a chemical system (atoms, chemical bonds, molecular substructures, molecules, and chemical reactions) can be represented in chemical hypergraph.

- *Natural*: Atoms are represented as nodes; all others (chemical bonds, molecular substructures, molecules, and chemical reactions) are represented as hyperedges because these are interactions and correlations in nature.
- *Robust*: Atoms and molecules have no inherent ordering within a chemical system. Chemical hypergraph is invariant to permutations on atoms and molecules within a given chemical system.
- *Interpretable*: It is important for chemists to understand the reasoning behind a molecule/reaction-level prediction. Chemical hypergraph provides means to interpret molecular or chemical reaction predictions, e.g., by capturing the correlation between the presence and absence of certain functional groups in molecules.
- *General*: Chemical hypergraph is agnostic to machine learning tasks and algorithms.

## 3.2 Related Work

We mention *Rxn Hypergraph* [16] here because it demonstrates the value of integrating molecular representation and chemical reaction representation in a single mathematical structure, which solves at once the chemical reaction representation and property-prediction problems, as the case with chemical hypergraph.

However, we should emphasize that, despite its name, Rxn Hypergraph does not use hypergraph with hyperedges representing high-order correlations. Instead, it uses traditional graph with pairwise edges. Therefore we put "hypergraph" in quotes here, and the same for *"hypernode"*.

Rxn Hypergraph exploits graph representations of molecular structures to develop and test a "hypergraph" attention neural network approach to solve at once the chemical reaction representation and property-prediction problems, At the molecular graph level, a chemical reaction with N reactants and M products is described by two distinct sets of disconnected molecular graphs R and P, with R representing the set of reactant molecules and P representing the set of product molecules. Each molecule graph G has a set of nodes (atoms) and a set of edges (bonds). Nodes have atom types as features, and edges have bond types as features.

The idea behind the Rxn Hypergraph of a chemical reaction is to efficiently construct message passing routes between these disconnected molecular graphs and form one connected "hypergraph" to represent the entire chemical reaction. To form this "hypergraph":

- For each of the disconnected molecular graph $G_i$, add a "hypernode" to the "hypergraph" as a mol-hypernode, representing a reactant or product molecule.

- Add a set of bidirectional edges connecting every pair of mol-hypernodes on either side of the reaction, mol-mol.
- Augment the "hypergraph" by adding two more "hypernodes" as rxn-hypernodes $x^r$ and $x^p$, one for the reactant and one for the product side of the reaction.
- Add a new type of edge to the "hypergraph": a set of unidirectional edges from each mol-hypernode to the rxn-hypernode of the same side of the reaction, mol-rxn.
- Add a set of bidirectional edges connecting every atom to the mol-hypernode of their parent molecule, molatom. Note that these edges connect the chemical reaction part of the "hypergraph" with the disconnected reactant and product molecular graphs.

As can be seen from above, Rxn Hypergraph is adhoc and complicated. *Chemical hypergraph*, on the contrary, is methodical and straightforward.

# 4 Molecular Hypergraph

For many applications, such as molecular property prediction, it is sufficient to consider only *molecules*, with their constituent atoms and chemical bonds but without their associated chemical reactions. In which case, *molecular hypergraph*, a simplified version of chemical hypergraph with only *simple hyperedges*, can be used to represent molecules. It allows machine learning models to, e.g.:

- Discover molecules with desired properties
- Develop molecular similarity measures

## 4.1 Molecular Hypergraph as Simple Hypergraph

A *molecular hypergraph* is a *simple hypergraph*, corresponding to the molecule hyperedge in chemical hypergraph (see Section 3.1 Chemical Hypergraph as Multilevel Hypergraph). *Nodes* in a molecular hypergraph represent *atoms*, and the *simple hyperedges* represent *chemical bonds / molecular substructures*. A molecular hypergraph represents molecules in a way that captures their chemical structures more accurately than traditional molecular graphs, allowing for the representation of *multicenter bonds* and *molecular substructures* like rings, delocalized bonds, conjugated bonds, and functional groups.

Here are *key features* of molecular hypergraphs:

- *Nodes*: Each node represents an *atom* in the molecule, with features such as atom type.

- *Bond / Substructure Hyperedges*: Each *simple hyperedge* represents a chemical bond or molecular substructure among atoms, with features such as bond / substructure type.

As an example, the molecule *benzene ($C_6H_6$)* contains 12 atoms, six C-H σ bonds, six C-C σ bonds, and one large delocalized π bond. Its molecular hypergraph consists of twelve nodes, twelve 2-order hyperedges, and one 6-order hyperedge, as visualized below:

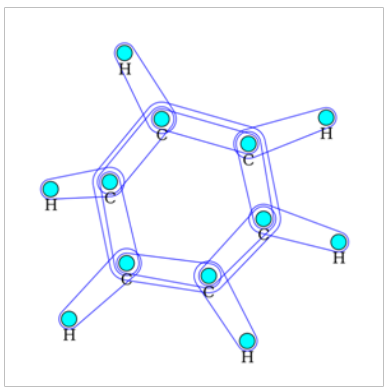

Molecular hypergraph: $C_6H_6$

## 4.2 Related Work

We mention *Molecular Hypergraph Neural Network (MHNN)* [17] here because it demonstrates the use of molecular hypergraph in deep learning for molecular property prediction. MHNN uses conventional undirected hypergraph, which is identical to simple hypergraph, to represent molecules.

MHNN is a deep learning model specifically designed to operate on molecular hypergraphs. It can learn from the rich information encoded in the hypergraph structure, leading to better property predictions. MHNN is based on bipartite representations of hypergraphs, which can efficiently operate on hypergraphs with hyperedges of various orders. The molecular hypergraph is initially transformed into an equivalent bipartite graph, wherein two distinct sets of vertices denote the nodes and hyperedges of the molecular hypergraph, respectively.

The message passing of MHNN relies on the bipartite representations converted from molecular hypergraphs. MHNN consists of multiple layers. For each layer:

1. Send messages from nodes to hyperedges

2. Update hyperedge embeddings
3. Send messages from hyperedges to nodes
4. Update node embeddings

The hypergraph embedding from nodes is obtained by summing all node embeddings, and the hypergraph embedding from hyperedges is obtained by summing all hyperedge embeddings. These are then used as inputs to a MLP for generating the prediction of MHNN.

MHNN demonstrates many benefits of using *molecular hypergraph* for deep learning:

- Capturing complex relationships: Molecular hypergraph can capture the intricate interactions between atoms in molecules, such as conjugated bonds, leading to more accurate property predictions.
- Data efficiency: Molecular hypergraph can achieve good performance even with limited training data.
- 3D structure independence: Molecular hypergraph might not require explicit 3D structural information, which can be computationally expensive to obtain..

## 5 Chemical Reaction Hypergraph

For many applications, such as chemical reaction network analysis, it is sufficient to focus on *chemical reactions* and treat their reactant and product molecules as opaque. In which case, *chemical reaction hypergraph*, a simplified version of chemical hypergraph with *directed hyperedges* representing chemical reactions and *nodes* representing molecules, can be used. It allows machine learning models to, e.g.:

- Predict reaction products and yields
- Identify feasible reaction pathways

### 5.1 Chemical Reaction Hypergraph as Directed Hypergraph

A *chemical reaction hypergraph* is a *directed hypergraph*, corresponding to the chemical reaction hyperedge in chemical hypergraph (see Section 3.1 Chemical Hypergraph as Multilevel Hypergraph), except that in this case both reactant (source) and product (target) hyperedges are *simple hyperedge*. *Nodes* in a chemical reaction hypergraph represent *molecules*, and the *directed hyperedge* represent *chemical reaction*. Each directed hyperedge contains, and specifies, the *reactant (source) molecules* and *product (target) molecules* of the chemical reaction.

Here are *key features* of chemical reaction hypergraphs:

- *Nodes*: Each node represents a *molecule* involved in the chemical reactions, with features such as molecular formula. Molecules can be reactants, products or catalysis.
- *Reactant and Product Hyperedges*: Each *simple hyperedge* represents the reactants or products of a chemical reaction. The hyperedge is *nested* within a chemical reaction hyperedge.
- *Chemical Reaction Hyperedge*: Each *directed hyperedge* represents a chemical reaction which involves multiple molecules. The hyperedge specifies the *source* and *target hyperedges,* which respectively represent the *reactant* and *product hyperedges* of the chemical reaction.

As an example, the reaction A + B → C + D is represented as the directed hyperedge {A, B} → {C, D} [9] shown below:

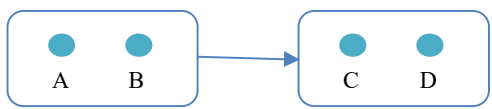

Chemical reaction hypergraph: A + B → C + D

## 5.2 Chemical Reaction Networks

*Chemical reaction networks (CRNs)* [22-23], defined by sets of *species* and possible *reactions* between them, are widely used to study *chemical systems*. Here is an *example CRN* [22] which we use for illustration:

- r1:  A $\rightleftarrows$ 2B
- r2:  A + C $\rightleftarrows$ D
- r3:  D → B + E
- r4:  B + E → A + C

A, B, C, D and E are chemical *species* (e.g., molecules). The objects that sit at the heads and the tails of the *reaction arrows*—A, 2B, A + C, D, and B + E—are called the *complexes* of the CRN. No complex reacts to itself and no complex is isolated—that is, each complex lies either at the head or at the tail of some reaction arrow.

Formally, a *CRN* consists of three sets [22]:

- a *finite set S*, elements of which are the *species (types)* of the CRN

- a *finite set C*, elements of which are the *complexes* of the CRN

- a set $\mathcal{R} \subset C \times C$ such that

    o for each $y \in C$, $(y, y) \notin \mathcal{R}$

    o for each $y \in C$ there is a $y' \in C$ such that $(y, y') \in \mathcal{R}$ or $(y', y) \in \mathcal{R}$

Members of $\mathcal{R}$ are the *reactions* of the CRN. For each (y, y') in $\mathcal{R}$, we say that complex y reacts to complex y', i.e., y → y'. y is called the *reactant complex* and y' is called the *product complex*. The component $y_s$ (corresponding to species s) of the complex y is called the *stoichiometric coefficient* of species s in complex y.

For the example CRN, we have

- $S$ = {A, B, C, D, E}

- $C$ = {A, 2B, A + C, D, B + E}

- $\mathcal{R}$ = {A → 2B, 2B → A, A + C → D, D → A + C, D → B + E, B + E → A + C}

In the complexes, the stoichiometric coefficients are:

- *A*: A=1, B=0, C=0, D=0, E=0

- *2B*: A=0, B=2, C=0, D=0, E=0

- *A + C*: A=1, B=0, C=1, D=0, E=0

- *D*: A=0, B=0, C=0, D=1, E=0

- *B + E*: A=0, B=1, C=0, D=0, E=1

Stoichiometric coefficients can be represented in matrix form, the so-called *stoichiometric matrix* **S**, with *species* as rows and *complexes* as columns. For the example CRN, **S** (written as row of column vectors) is:

**S** = ((1; 0; 0; 0; 0), (0; 2; 0; 0; 0), (1; 0; 1; 0; 0), (0; 0; 0; 1; 0), (0; 1; 0; 0; 1))

*Chemical reaction hypergraph* provides natural representation of CRNs. Specifically, for a CRN we have

- *Nodes*: Each node represents a *species (type)* in the CRN.
- *Complex Hyperedge*: Each *simple hyperedge* represents a *complex* (*reactants* or *products*) in the CRN.
- *Chemical Reaction Hyperedge*: Each *directed hyperedge* represents a chemical reaction in the CRN.

Since a node represents a species type, instead of a species individual (there may be more than one for a given type), the incidence matrix $H$ of a complex hyperedge or chemical reaction hyperedge is denoted by the *stoichiometric matrix $S$*, with *species (types)* as rows and *complexes* or *chemical reactions* as columns.

An example of chemical reaction hypergraphs for CRN is provided below.

## 5.3 Related Work and Example

We mention *Directed and Weighted Metabolic Hypergraphs (DWMH)* [18] here because it demonstrates the use of chemical reaction hypergraph in chemical reaction network analysis. DWHN uses conventional directed hypergraph, which is essentially identical to directed hypergraph, to represent chemical reactions.

A *metabolic network* is a highly organized system of chemical reactions that occur in living organisms to sustain life and regulate cellular processes. Metabolic networks are incredibly complex because of the large number of reactions and the intricate web of interactions between molecules. Several reactions are involved in metabolism, grouped into various metabolic pathways. A metabolic pathway is an ordered chain of reactions in which metabolites are converted into other metabolites or energy. Here is an example of a small metabolic network composed of three reactions (the first being reversible) and five metabolites:

- r1:  3a + 2b <-> c
- r2:  c + b -> a + 4d
- r3:  d -> 2e

DWMH proposes *metabolic hypergraph*, a directed hypergraph with hyperedge-dependent node weights, as a framework to represent metabolic networks. This hypergraph based representation captures high-order interactions among metabolites and reactions, as well as the directionalities of reactions and stoichiometric weights, preserving all essential information.

DWMH further proposes the communicability and the search information as metrics to quantify the robustness and complexity of directed hypergraphs.

In terms of *chemical reaction hypergraph*, the metabolic hypergraph for the example metabolic network can be visualized as:

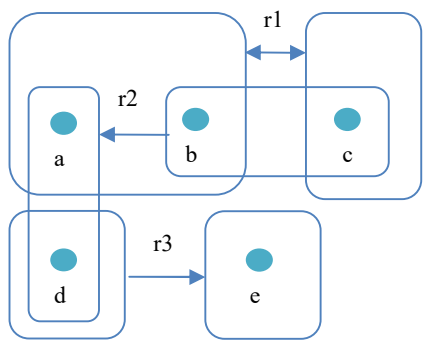

In the above, the *stoichiometric coefficients* are:

- r1+: a = -3, b = -2, c = 1, d = 0, e = 0
- r1-: a = 3, b = 2, c = -1, d = 0, e = 0
- r2: a = 1, b = -1, c = -1, d = 4, e = 0
- r3: a = 0, b = 0, c = 0, d = -1, e = 2

Note that the so-called hyperedge-dependent stoichiometric node weights used in [18] are not needed because they are simply stoichiometric coefficients (i.e., stoichiometric matrix **S**, in place of incidence matrix **H**).

## 6 Conclusion

In this work, we make two original and substantial contributions:

- We propose a new definition of hypergraph that unifies the concepts of undirected, directed and nested hypergraphs, and that is uniform in using hyperedge as a single construct for representing high-order correlations among things, i.e., nodes and hyperedges. Specifically, we define a hyperedge to be a simple hyperedge (a node set), a nesting hyperedge (a hyperedge set), or a directed hyperedge (a nesting hyperedge with the notion of direction applied between member hyperedges). The new definition resolves two major issues with the conventional definition of hypergraph: (1) there is not a standard definition of directed hypergraph, and the notion of direction may be applied either between hyperedges or between node sets within the same hyperedge, and (2) there is not a formal definition of nested hypergraph, in which hyperedges can contain other hyperedges. The uniformity and power of this new definition, with visualization, should facilitate the use of hypergraphs for representing (hierarchical) high-order correlations in general and chemical systems in particular.

- We propose the use of chemical hypergraph, a multilevel hypergraph (with simple, nesting and directed hyperedges), as a single mathematical structure for representing chemical systems consisting of molecules, their constituent atoms and chemical bonds, and their chemical reactions. Simplified versions of chemical hypergraph can be used to represent partial chemical systems, such as molecules or chemical reactions, in the form of molecular hypergraph or chemical reaction hypergraph respectively. Chemical hypergraph provides a uniform, powerful and flexible representation of chemical systems (from atoms, to chemical bonds / molecular substructures, to molecules, to chemical reactions) for machine learning and other usage. It is complete, natural, robust, interpretable and general.

Furthermore, we demonstrate the validity and applicability of the new unified and uniform definition of hypergraph in non-chemical domains in the appendix.

**Acknowledgement:** Thanks to my wife Hedy (郑期芳) for her support.

# Appendix. Applications in Non-chemical Domains

To demonstrate the validity and applicability of the new unified and uniform definition of hypergraph in non-chemical domains, we apply it to two publicly available example hypergraph applications in the following.

## A.1 Analysis of the Novel Les Miserables

An analysis of the high-order interactions among characters in the novel Les Miserables using hypergraph is provided in *HyperNetX (HNX)* [19-20] as a tutorial. The novel is broken into five parts, which are referenced as *volumes*: Fantine**,** Cosette**,** Marius**,** St. Denis, and Jean Valjean. Each volume is subdivided into *books*, each book into *chapters*, and each chapter into *scenes*. The novel involves 80 *characters*. HNX supports only *simple hypergraph*. As a result, numerous hypergraphs at different level and granularity are created in the tutorial for analysis of the novel:

- A hypergraph with *volumes* as hyperedges and *characters* as nodes.
- Five hypergraphs, each for a volume, with *books* as hyperedges and *characters* as nodes.
- Five hypergraphs, each for a volume, with *scenes* as hyperedges and *characters* as nodes.
- For Book 1 in Volume 1, a hypergraph with *scenes* as hyperedges and *characters* as nodes.

The above list of hypergraphs, though many, is far from being exhaustive of all possible levels and granularity.

We redesign the analysis by using the *new unified and uniform definition* of hypergraph, which supports *simple, nesting and directed hyperedges*. We find that the high-order interactions among characters in the novel can be fully represented by a single *multilevel hypergraph*, which reflects the structure and granularity, as well as their ordering, of the novel. As such, the representation is simpler, complete and more powerful. The multilevel hypergraph consists of:

- *Nodes*: Each node represents a *character* in the novel.
- *Scene Hyperedges*: Each *simple hyperedge* represents a scene with characters. The hyperedge is *nested* within a *chapter hyperedge* because the scene exists as part of a chapter.
- *Chapter Hyperedges*: Each *nesting hyperedge* represents a chapter which consists of multiple scenes. The hyperedge is *nested* within a *book hyperedge* because the chapter exists as part of a book.
- *Book Hyperedges*: Each *nesting hyperedge* represents a book which consists of multiple chapters. The hyperedge is *nested* within a *volume hyperedge* because the book exists as part of a volume.
- *Volume Hyperedges*: Each *nesting hyperedge* represents a volume which consists of multiple books.
- *Directed Hyperedges*: Each *directed hyperedge* represents an ordering among parts of the novel at the same level and granularity: *volumes*, *books* (within the same volume), *chapters* (within the same book), and *scenes* (within the same chapter). The hyperedge specifies the *source and target hyperedges* at the same level and granularity.

This is visualized below for the multilevel nesting:

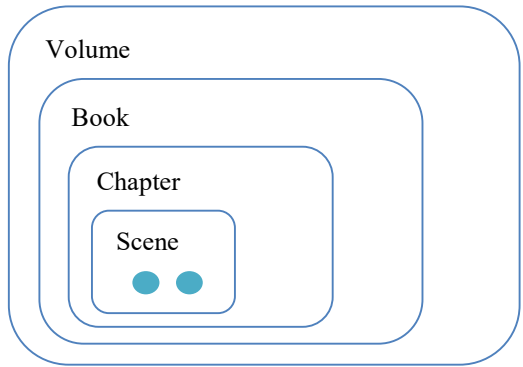

and for the *ordering* among volumes:

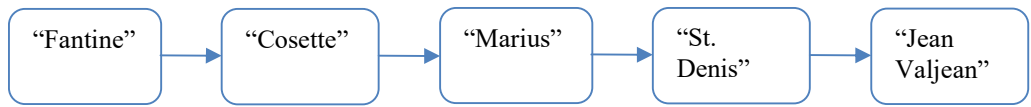

## A.2 Representation of Clinical Notes

A *taxonomy-aware multi-level hypergraph* for representation of *clinical notes* is proposed in *TM-HGNN* [21], where multi-level hypergraphs are used to assemble useful *neutral words* with rare *keywords* via *note and taxonomy level hyperedges* to retain the clinical semantic information.

TM-HGNN supports only *simple hypergraph*. As such, only note-level and taxonomy-level hyperedges are used for representation of clinical notes:

- *Nodes*: Each node represents a *word* in clinical notes or taxonomies.
- *Note Hyperedges*: Each *simple hyperedge* represents a clinical note consisting of words.
- *Taxonomy Hyperedges*: Each *simple hyperedge* represents a clinical taxonomy described by words.

Taxonomy hyperedges may be implicitly nested.

We extend the representation of clinical notes by using the *new unified and uniform definition* of hypergraph, which supports *simple, nesting and directed hyperedges*. We find that the clinical semantic information provided by clinical notes can be extensively represented by a *hierarchical multilevel hypergraph*, which reflects the content, semantics and context of clinical notes. As a result, the representation is more complete and powerful. The hierarchical multilevel hypergraph consists of:

- *Nodes*: Each node represents a *word* in clinical notes or taxonomies.
- *Note Hyperedges*: Each *simple hyperedge* represents a clinical note consisting of words. The hyperedge is *nested* within a *patient hyperedge* and a *physician hyperedge* because the note is associated with a patient and written by a physician.
- *Taxonomy Hyperedges*: Each *simple hyperedge* represents a clinical taxonomy described by words. The hyperedge may be *nested* within a *parent taxonomy hyperedge*.
- *Patient Hyperedges*: Each *nesting hyperedge* represents a patient who has associated clinical notes.
- *Physician Hyperedges*: Each *nesting hyperedge* represents a physician who has written clinical notes.
- *Parent Taxonomy Hyperedges*: Each *nesting hyperedge* represents a parent taxonomy. The hyperedge may be *nested* within a *parent taxonomy hyperedge*.
- *Directed Hyperedges*: Each *directed hyperedge* represents a *patient to physician* relationship or a *physician to patient* relationship. The hyperedge specifies the *source* and *target hyperedges*.

This is visualized below for the multilevel nesting:

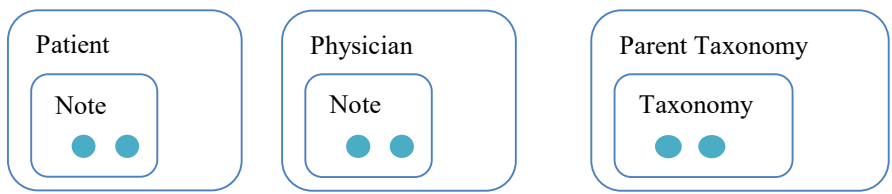

and for the *patient-physician* relationships:

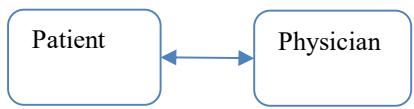